\documentclass[conference]{IEEEtran}
\IEEEoverridecommandlockouts
\usepackage[bookmarks=false]{hyperref}
\usepackage{amsmath,amsfonts,amsthm} 
\usepackage{url}
\usepackage{bm}
\usepackage{xfrac}
\usepackage{stfloats}
\usepackage{fixltx2e}
\usepackage{booktabs}
\usepackage[caption=false,font=footnotesize]{subfig}
\usepackage{array}
\usepackage{algorithm}
\usepackage{multirow}
\usepackage{cite}
\usepackage{wrapfig}
\usepackage{lscape}
\usepackage{rotating}
\usepackage{epstopdf}
\usepackage{float}
\usepackage{algpseudocode}
\usepackage{comment}
\usepackage{stfloats}

\usepackage[normalem]{ulem}
\useunder{\uline}{\ul}{}

\usepackage{eqparbox,array}

\renewcommand{\algorithmiccomment}[1]{\bgroup\hfill\scriptsize//~#1\egroup}

\usepackage{tikz}
\usetikzlibrary{decorations.pathreplacing,calc}

\newcommand\Train{\mathcal{T}}

\begin{document}
\algnewcommand{\algorithmicgoto}{\textbf{go to}}%
\algnewcommand{\Goto}[1]{\algorithmicgoto~\ref{#1}}%
\algnewcommand\algorithmicinput{\textbf{Input:}}
\algnewcommand\Input{\item[\algorithmicinput]}
\algnewcommand\algorithmicoutput{\textbf{Output:}}
\algnewcommand\Output{\item[\algorithmicoutput]}

\title{Multi-label learning for dynamic model type recommendation}

\author{\IEEEauthorblockN{Mariana A. Souza\IEEEauthorrefmark{1},
Robert Sabourin\IEEEauthorrefmark{1},
George D. C. Cavalcanti\IEEEauthorrefmark{2} and
Rafael M. O. Cruz\IEEEauthorrefmark{1}
}
\IEEEauthorblockA{\IEEEauthorblockA{\IEEEauthorrefmark{1}\'{E}cole de Technologie Sup\'{e}rieure - Universit\'{e} du Qu\'{e}bec, Montreal, Quebec, Canada\\
Email: mariana.araujo.souza@gmail.com, robert.sabourin@etsmtl.ca, rafaelmenelau@gmail.com}
\IEEEauthorrefmark{2}Centro de Inform\'{a}tica - Universidade Federal de Pernambuco, Recife, Pernambuco, Brazil\\
Email: gdcc@cin.ufpe.br}}

\maketitle

\begin{abstract}
Dynamic selection techniques aim at selecting the local experts around each test sample in particular for performing its classification. 
While generating the classifier on a local scope may make it easier for singling out the locally competent ones, as in the online local pool (OLP) technique, using the same base-classifier model in uneven distributions may restrict the local level of competence, since each region may have a data distribution that favors one model over the others. 
Thus, we propose in this work a problem-independent dynamic base-classifier model recommendation for the OLP technique, which uses information regarding the behavior of a portfolio of models over the samples of different problems to recommend one (or several) of them in a per-instance manner. 
Our proposed framework builds a multi-label meta-classifier responsible for recommending a set of relevant base-classifier models based on the local data complexity of the region surrounding each test sample. 
The OLP technique then produces a local pool with the model that yields the highest probability score of the meta-classifier. 
Experimental results show that different data distributions favored different model types on a local scope. 
Moreover, based on the performance of an ideal model type selector, it was observed that there is a clear advantage in choosing a relevant base-classifier model for each test instance in particular. 
Overall, the proposed model type recommender system yielded a statistically similar performance to the original OLP with fixed base-classifier model. 
However, the proposed framework struggled to recommend at least one relevant model type specially for the samples with low labelset cardinality. 
Given the novelty of the approach and the gap in performance between the proposed framework and the ideal selector, we regard this as a promising research direction. 
\\
Code available at \url{github.com/marianaasouza/dynamic-model-recommender}.
\end{abstract}

\section{Introduction}

\par Multiple Classifier Systems (MCS) combine the responses of several classifiers in the hopes that the combined system outperforms each individual base-classifier \cite{onCombin,wozniak2014survey}. 
MCS are usually divided into three phases \cite{cruz_dynamic_2018}: generation, in which the base-classifiers from the pool are generated, selection, in which a subset of the classifiers may be selected to perform the classification, and aggregation, in which the responses of the selected base-classifiers are combined. 
The classifier selection may be either static or dynamic, with the former being performed during training and the latter during generalization. 
The reasoning behind dynamic selection techniques is that each classifier in the pool may be a local expert in different regions of the feature space, so the dynamic selection schemes aim at selecting the base-classifier(s) that are best fit for labelling each test instance in particular. 
Dynamic selection techniques were shown to outperform static selection approaches specially on ill-defined problems \cite{DESsurvey}. 

\par Yet since most pool generation methods used in dynamic selection schemes are classical techniques designed for static selection \cite{cruz_dynamic_2018}, such as Bagging \cite{bagging} and Boosting \cite{boosting}, they generate the base-classifiers with a global perspective of the problem, so producing a local expert in the vicinity of all test instances is not guaranteed. 
Dynamic selection techniques were also shown to have difficulty in selecting a locally competent classifier even when it exists in the pool \cite{oliveira2017online,mariana}. 
In a previous work \cite{souza2019online}, it was proposed an online local pool generation method (OLP). 
The OLP produces hyperplanes on the fly in the area surrounding each test instance near class borders, so as to guarantee the presence of locally accurate classifiers in the region and thus facilitate their selection by the dynamic selection techniques. 
Using the locally generated pool was shown to increase the frequency at which the most competent classifier is selected by the evaluated dynamic selection schemes in comparison to using a globally generated pool \cite{souza2019online}, and also to work well over imbalanced problems \cite{souza2019oneval}. 

\par However, the production of local experts by the OLP technique was restricted by the base-classifier model used in the pool, which in this case are only two class Perceptrons. 
To the best of our knowledge, the choice of base-classifier model used in pool generation techniques is always done a priori in the literature regarding ensemble methods.  
However, in uneven distributions, each local region in the feature space may have different characteristics, such as data topology, class balance, class overlap and data density, among others. 
Since the idea is to select the base-classifier(s) that are experts in a given region, it would follow that the best model to learn the data from that region depends on its local data distribution. 
For instance, using the Perceptron as base-classifier may be far from ideal when trying to label a query sample located near a non-linearly separable local class border, while it would make sense to use it when labelling an instance near a linearly separable one. 
So, our hypothesis is that, by choosing a relevant base-classifier model for each instance in particular, we may be able to produce more locally competent base-classifiers in that region, and thus yield an improvement in performance compared to using a fixed base-classifier model for all samples.

\par Thus, we propose a framework for base-classifier model recommendation in order to, given a local data distribution, indicate which model is most suitable to be used for each test sample, with the purpose of yielding a pool of local experts for the online local pool scheme. 
We wish to answer the following research questions with our model type recommender system: (a) which data characteristics affect the performance of each base-classifier model on a local scope?, and (b) can we use this information to recommend a suitable model for each given query sample and improve the online method's performance?
To do so, we use a problem-independent multi-label meta-classifier, which is obtained using information from the behavior of the base-classifier models over instances from previous datasets. 
We construct the multi-label set of the meta-problem using the class probabilities of each base-classifier model over several problems. 
We then associate its responses for each sample to its meta-features, which are comprised of local complexity measures obtained over the neighborhood of the corresponding sample in the training set, and train the multi-label meta-classifier afterwards. 
In generalization, the meta-features of each query instance are first extracted using its neighboring samples in the training set, which are then fed into the meta-classifier, who outputs the relevant base-classifier models for that instance. 

\par Our proposed framework is closely related to the algorithm recommendation area. 
Several recommender systems for algorithm selection based on data complexity measures were proposed in the literature. 
In \cite{garcia2018classifier}, the authors use a meta-regressor, built with the extracted data complexity measures of several datasets, to select the best-performing classification model for a given unknown problem. 
A somewhat similar framework is proposed in \cite{deng2018automatic}, in which the authors propose a set of meta-features based on data complexity extracted from a kernel matrix generated from the data, in order to recommend the most suitable classification model using an NN rule. 
Another algorithm recommender system was proposed in \cite{das2016meta} for fault prediction in software projects. 
The meta-data was obtained by extracting a set of meta-features, including simple, statistical and data complexity measures, from the data of previous software projects and assigning which model in the portfolio yielded the best performance, according to the balance criterion \cite{menzies2006data}, over each project as the meta-label. 
However, none of these recommender systems base their model recommendation procedure on the local data characteristics within a given dataset. 
Moreover, they recommend the models for an entire problem, not for each instance in particular. 
To the best of our knowledge, \textit{per-sample} meta-learning for algorithm recommendation was only explored in certain fields, such as combinatorial search problems \cite{kotthoff2016algorithm} and collaborative filtering 
\cite{collins2018novel}.
%
%

\par This work is divided as follows. 
Section \ref{sec:proposed-framework} describes our proposed base-classifier model recommender system. 
Experimental results are presented in Section \ref{sec:experimental-results}. 
Lastly, we present our concluding remarks in Section \ref{sec:conclusion}. 

\section{Proposed framework}
\label{sec:proposed-framework}

\subsection{Online local pool method} 
\label{sec:olp}

\par Before delving into the proposed model type recommender system, we briefly present the OLP technique next.
The OLP technique attempts to exploit the properties of the Oracle \cite{oracle} on a local scope in order to guide the generation of the local base-classifiers. 
The Oracle is an ideal selector which always chooses the base-classifier in the pool that labels a given test sample correctly, if such classifier exists. 
The OLP technique was shown to yield a similar performance to state-of-the-art ensemble methods \cite{souza2019online}, and also to perform quite well on imbalanced distributions \cite{souza2019oneval}.


\par In the offline phase of the OLP, the  K-Disagreeing Neighbors (KDN) \cite{smith2014instance} estimates of each training sample is computed, in order to identify which areas of the feature space present a local class border. 
The KDN measure calculates the proportion of samples from a different class in the neighborhood of a given sample.



\par
\begin{figure*}
    \centering
    \includegraphics[width=0.78\textwidth]{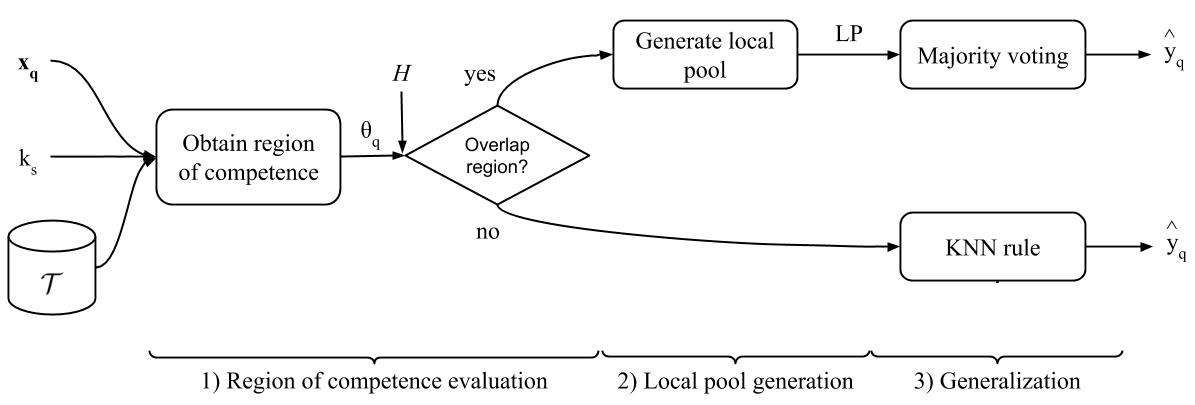}
    \caption{Overview of the online phase of the online local pool generation method (from \cite{souza2019oneval}).
    The symbols $k_s$, $H$ and LP mean the dynamic selection technique's neighborhood size, the set of KDN estimates, and the local pool, respectively. 
    }
    \label{fig:baseline-on}
\end{figure*}

 The online phase of the OLP is described in Figure \ref{fig:baseline-on}, in which $k_s$ is the dynamic selection neighborhood size, $H$ is the set of KDN estimates and $LP$ is the local pool. 
It is divided in three steps: region of competence estimation, local pool generation and generalization. 
In the first step, the neighborhood $\theta_{q}$ of the query sample is first obtained using regular K-NN, with size $k_s$ over the training set, and then evaluated based on the KDN scores stored in $H$, obtained in the offline phase. 
If none of the sample's neighbors are borderline samples, that is, if their KDN score is zero, then the procedure goes directly to the third and last step, generalization, and the K-NN classifier yields the predicted label $\hat{y}_q$.  
If, however, any of the neighbors $\mathbf{x_i} \in \theta_q$ is a borderline sample, the region is identified as a borderline area and the local pool (LP) is generated in the next step.

\par
\begin{figure*}[!htb]
		\centerline{		
		\includegraphics[width=0.78\textwidth]{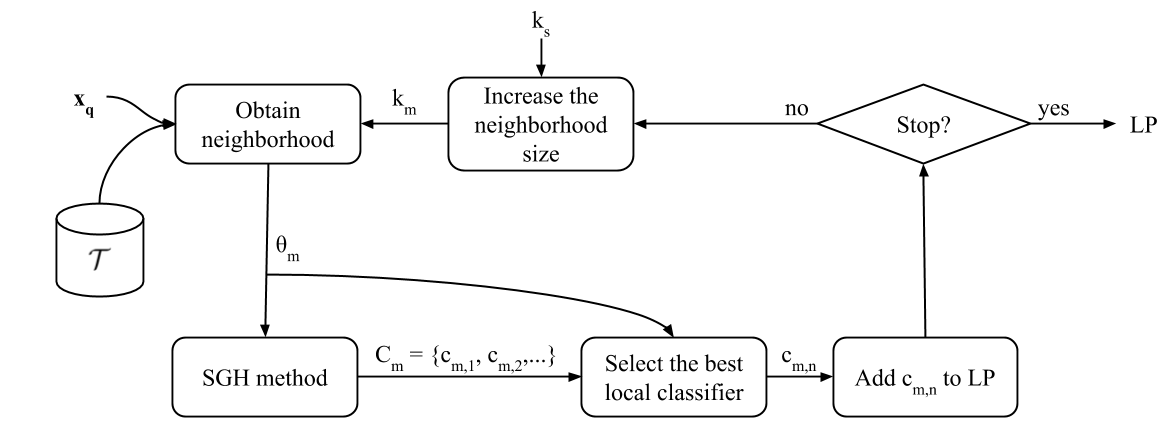}}
		\caption{Local pool generation step (from \cite{souza2019oneval}). 
		The symbols $k_s$ and LP mean the dynamic selection technique's neighborhood size and the local pool, respectively.
		}
		\label{fig:gen-lp}
\end{figure*}

 In the second step, the LP is produced iteratively, and in each iteration the most locally competent classifier produced in that iteration is added to the final pool (Figure \ref{fig:gen-lp}). 
In a given \textit{m}-th iteration, the query sample's neighboring instances in the training set $\Train$ are obtained using a neighborhood size of $k_{m}$, calculated based on $k_s$. 
For two-class problems, the K-Nearest Neighbors Equality (K-NNE) \cite{knne}, which selects the same amount of neighbors from each class, is used in this step. 

\par The query sample's neighborhood $\theta_{m}$ is then used as input to the Self-generating Hyperplanes (SGH) method \cite{mariana}, a pool generation method that yields an Oracle accuracy rate of 100\% over the input dataset. 
The SGH then produces a local subpool $C_{m}$ in which the presence of at least one competent classifier $c_{m,k} \in C_{m}$ for each instance in $\theta_{m}$ is guaranteed. 
The indexes in the classifiers' notation indicates that the classifier $c_{m,k}$ is the \textit{k}-th classifier from the \textit{m}-th subpool. 

\par Then, the most competent classifier $c_{m,n}$ from $C_{m}$ in the region delimited by the neighborhood $\theta_{q}$ is selected by a DCS technique and added to the local pool. 
The selection by a DCS technique is performed at this stage, and not after the LP is completed, because the subpool generation by the SGH method yields too diverse classifiers \cite{mariana}, so not only is it not fit for DES techniques, but also it may generate classifiers that are near opposite to the local border \cite{souza2019online}. 
Thus, the dynamic selection is performed concurrently with the generation. 
The same procedure using the SGH method is performed in iteration $m+1$ with the neighborhood size $k_{m+1}$ increased by 2, in order to not only adjust the locality of the classifiers but also to provide a different set of training samples and produce slightly diverse base-classifiers. 
This process is then repeated until the local pool contains a predefined amount ($M$) of locally accurate classifiers.

\par In the last step, generalization, the predicted label $\hat{y}_q$ of the query sample $\mathbf{x_q}$ is produced (Figure \ref{fig:baseline-on}). 
If the LP was generated, the responses of the base-classifiers in LP are combined using the majority voting rule. 
Otherwise, the K-NN is used to obtain $\hat{y}_q$.

\subsection{Dynamic model type recommendation system}
\label{sec:proposed-metaframework}

\par Though the OLP yielded promising results reported in previous works, it presents several limitations, one of the greatest being its local pool generation procedure based on the SGH method. 
Although the latter presents interesting qualities, the base-classifier generation is done using an heuristic and, for this reason, it can only produce two-class Perceptrons, yielding hyperplanes that are not always well adjusted to the border depending on the local data distribution around the query sample. 
Since different classification model types fit the data in different ways, our hypothesis is that locally training a suitable base-classifier model according to the local data distribution may be advantageous for producing a more competent set of classifiers for the OLP. 
Thus, we propose a dynamic base-classifier model recommender system which indicates, for each test instance, the \textit{relevant} base-classifier models considering the data complexity around the sample. 

The choice of a suitable base-classifier model to be trained for each instance can be formulated as a meta-learning problem, in which: 
\begin{itemize}
    \item The meta-classes correspond to the model types that are suitable for a particular instance $\mathbf{x_i}$. 
    \par Since more than one base-classifier model may be suitable for each given sample $\mathbf{x_i}$, it is associated with a meta-labelset $U_i$, or its corresponding meta-label vector $\mathbf{u_i}$, a binary vector indicating the relevant models.
    \item Each element $v_{i,j}$ of the meta-feature vector $\mathbf{v_i}$ corresponds to a different complexity measure extracted in the neighborhood of the sample $\mathbf{x_i}$.
    \item A multi-label meta-classifier is trained on the meta-dataset to predict which base-classifier models are relevant for a given query sample $\mathbf{x_q}$, according to its meta-feature vector $\mathbf{v_q}$ extracted in its neighborhood.
\end{itemize}

\subsection{Meta-training step}

\par
\begin{figure*}[]
    \centering
    \includegraphics[width=0.8\textwidth]{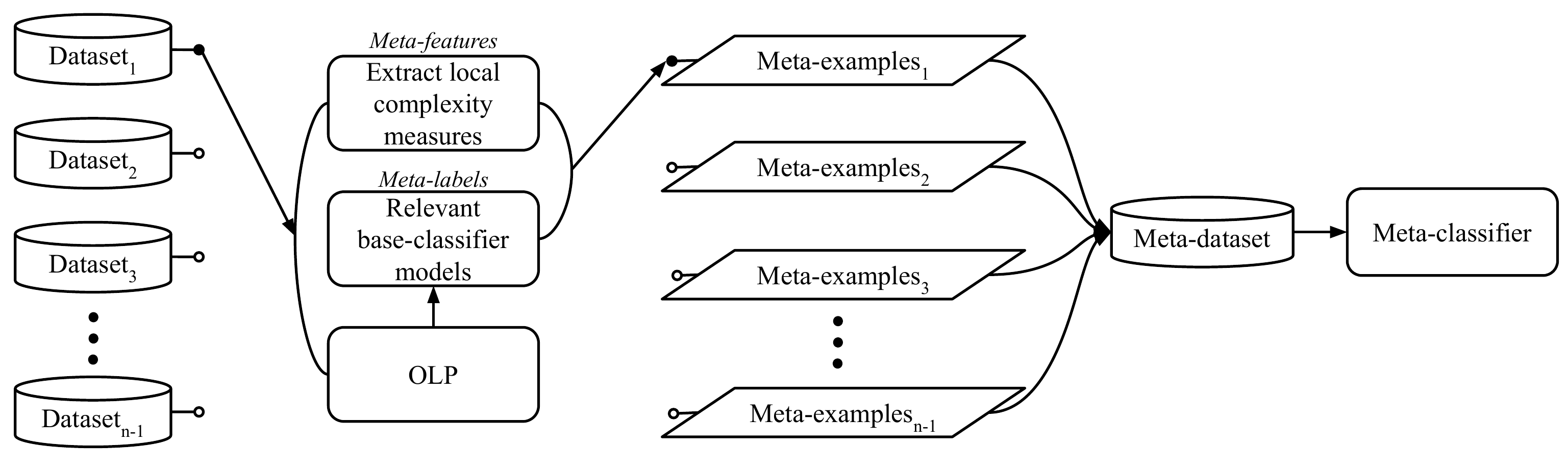}
    
    \caption{Overview of the training phase of the proposed framework.}
    \label{fig:overview-meta-training}
\end{figure*}

\begin{figure*}[]
    \centerline{
        \includegraphics[width=0.8\textwidth]{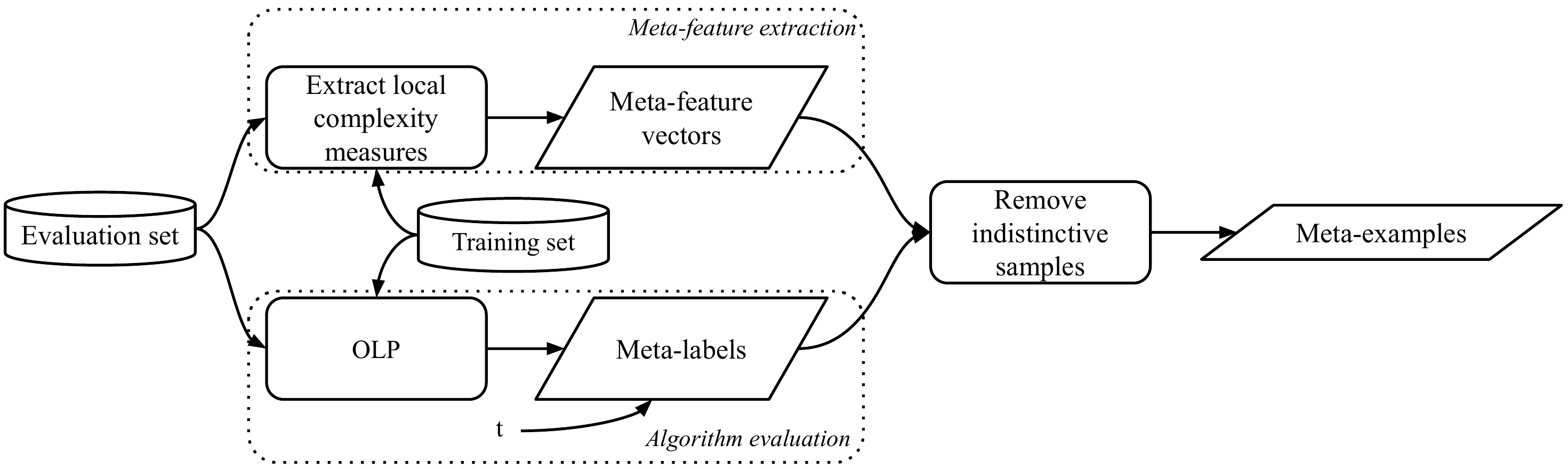}}
    \caption{Meta-feature extraction and algorithm evaluation for each dataset in the meta-training phase. The threshold $t$ is used to define the relevance of each model for a given sample.}
    \label{fig:training-ind-dataset}
\end{figure*}

 In the meta-training step (Figure \ref{fig:overview-meta-training}), we obtain our meta-data by evaluating the OLP technique with different model types and associating, for each instance, the information regarding which ones were successful (meta-labels) to the complexity measures extracted in the sample's neighborhood (meta-features). 
We then join the meta-examples obtained from several problems to form the meta-training set, which is used to train our meta-classifier. 
It is important to note that, since for each instance in a given problem there may be more than one suitable base-classifier model, our meta-learning problem is also a multi-label one. Thus, our meta-classifier must be able to deal with this scenario.
Figure \ref{fig:training-ind-dataset} shows in more detail the meta-feature extraction and the algorithm evaluation for each dataset from Figure \ref{fig:overview-meta-training} individually. 

\paragraph{Meta-feature extraction} 

\par In order to characterize the local data complexity of each sample in the evaluation set (Figure \ref{fig:training-ind-dataset}), we use 12 data complexity measures described in \cite{lorena2018complex}. 
Our reasoning for choosing these measures was based on the sort of information each of them brings for characterizing a local data distribution. 
We selected a subset of measures that cover all data complexity aspects described in \cite{lorena2018complex} with the exception of dimensionality, which is problem-dependent and would not help characterizing a local area in a problem-independent manner. 
Most of the chosen complexity measures were also shown to have a good discriminating power for predicting algorithm performance on a global scope \cite{garcia2018classifier, munoz2018instance}, specially the distance-based measures. 
The complexity measures selected as meta-features for the proposed framework are: 

\begin{itemize}
    \item Maximum Individual Feature Efficiency (F3): 
    This measure assesses the degree of ambiguity of the feature which presents the smallest overlap of values between samples from different classes. 
    
    \item Collective Feature Efficiency (F4): 
    The F4 measure gives an insight on the degree of efficiency provided by a given set of features, and is calculated using the F3 measure iteratively over a given set of points. 
    
    \item Error Rate of Linear Classifier (L2): 
    The L2 measure is defined as the error rate of a linear SVM trained over the input dataset. 
    
    \item Non-Linearity of a Linear Classifier (L3): 
    The L3 measure tries to quantify the degree of linearity of a problem's class borders, and is defined as the error rate of a linear classifier obtained with the original training set over prototypes generated via interpolation of the training points.
    
    \item Fraction of Borderline Points (N1): 
    The N1 measure is obtained by generating a minimum spanning tree (MST) using the distance matrix from all points of the input set and then calculating the proportion of samples that are connected to a sample from a different class, thus conveying the size and degree of complexity of the decision boundary.
    
    
    \item Ratio of Intra/Extra Class Nearest Neighbor Distance (N2): 
    The N2 measure estimates the iter/intra class relationship of the data by computing the ratio between the sum of the distances between the nearest neighbors (1-NN) of the same class and the sum of the distances between the nearest neighbors that possess different labels, for the entire input set. 

    
    \item Error Rate of the Nearest Neighbor Classifier (N3): 
    The N3 measure is defined as the error rate of the 1-NN classifier over the input dataset, computed using a leave-one-out procedure.
    
    \item Non-Linearity of the Nearest Neighbor Classifier (N4): 
    The N4 measure is similar to the L3 measure, in that it generates a few prototypes via interpolation 
    and evaluates a classifier, trained with the original training samples, over the newly generated prototypes. In the case of the N4, the classifier used is the 1-NN.
    
    \item Local Set Average Cardinality (LSC): 
    The LSC is defined as the average number of samples within the local set (LS) of each instance in the input set. 
    The LS of a given instance is the set of samples that share the same label as the target instance while being closer to it than its nearest enemy. 
    
    

    \item Average density of the network (Den):
    The Density measure is computed using a graph constructed so that each node is a training instance and each edge is a distance-based weighted connection between them, with the edges with distance greater than a given threshold are discarded, as well as all edges connecting samples from different classes. 
    The measure is then calculated as the normalized number of edges in the graph.
    
    
    \item Entropy of class proportions (C1):
     The C1 measure estimates the normalized entropy of the class sizes, giving an insight into the class imbalance of the data.

    \item Imbalance ratio (C2):
    Also referred to as IR, the imbalance ratio of binary problems is defined as the ratio between the number of samples in the majority class and the number of samples from the minority class. 
    
\end{itemize}

\par
\begin{figure}
    \centerline{
    \includegraphics[width=0.4\textwidth]{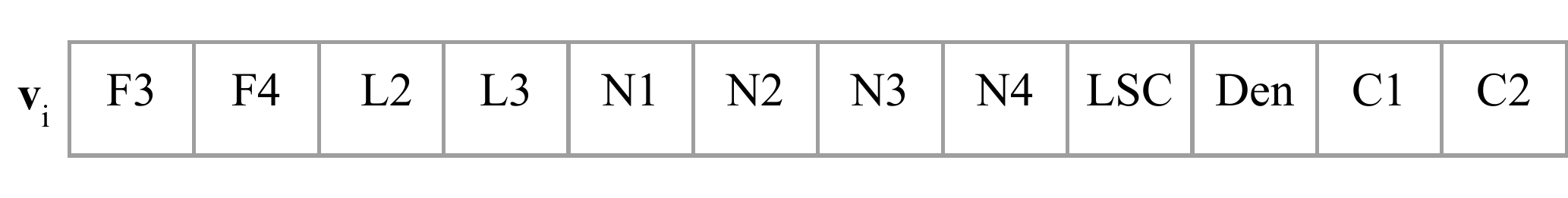}
    }
    \caption{Example of meta-feature vector.}
    \label{fig:ex-feature-vector}
\end{figure}

 In the meta-feature extraction step, we calculate for each sample $\mathbf{x_i}$ in the evaluation set (Figure \ref{fig:training-ind-dataset}) its neighborhood over the training set using the K-NN, with neighborhood size $k'$. 
We then extract the 12 complexity measures over the neighborhood of $\mathbf{x_i}$, yielding the meta-feature vector $\mathbf{v_i}$ depicted in Figure \ref{fig:ex-feature-vector}.

\paragraph{Algorithm evaluation} 

\par
\begin{figure}
    \centering
    \includegraphics[width=0.4\textwidth]{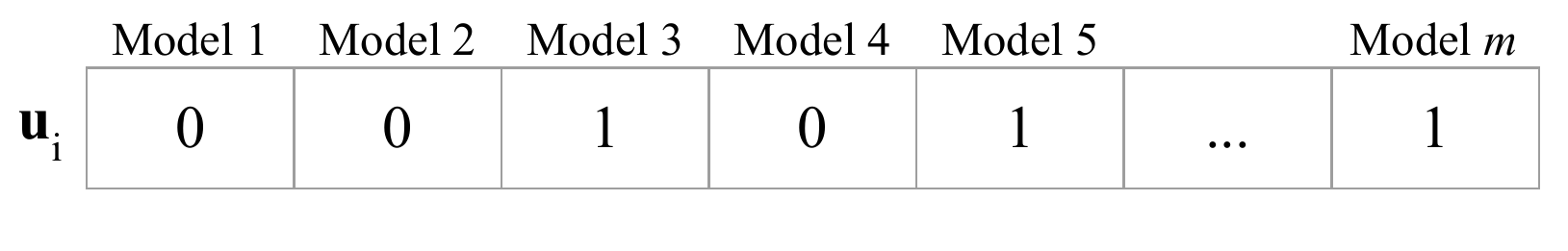}
    \caption{Example of meta-label vector. 
    The assigned value is $1$ if the model correctly classified the sample with output class probability above a threshold $t$, or $0$ otherwise.
    }
    \label{fig:ex-label-vector}
\end{figure}

 All samples in the evaluation set are tested using the OLP \textit{m} times, with \textit{m} being the number of model types considered in the meta-learning framework's portfolio. 
So, for each sample $\mathbf{x_i}$, the base-classifiers used in the executions that yielded the correct label $y_i$ with class probability above a threshold $t$ are referenced as relevant for that sample. 
The vector of meta-labels $\mathbf{u_i}$ corresponding to the sample $\mathbf{x_i}$ is illustrated in Figure \ref{fig:ex-label-vector}. 
Each column indicates the relevance of the base-classifier model for that sample. 
A relevant model for a given sample $\mathbf{x_i}$ is one with which the OLP technique was able to correctly label with output class probability above a threshold $t$. 
If the OLP with the base-classifier model was unable to classify the sample correctly, or the class probability was below $t$, the model is deemed non-relevant to that sample and its corresponding value in $\mathbf{u_i}$ is $0$. 
That way, we indicate to the meta-classifier which base-classifier models are indeed more likely to successfully learn the local data distribution. 

\par We then remove the samples for which all base-classifier models yielded the same response, referred to in Figure \ref{fig:training-ind-dataset} as \textit{indistinctive} samples. 
Since for these samples any model will produce the same output, they do not help in discriminating between the base-classifier models. 
Thus, similarly to \cite{cruz2015meta}, we remove the indistinctive samples in order for the meta-classifier to focus on the distinctive ones.

\paragraph{Meta-classifier training}

\par Lastly, the meta-classifier is trained with the meta-data. 
Since our meta-problem is a multi-label one (Figure \ref{fig:ex-label-vector}), we need a multi-label learning method for dealing with it. 
For simplicity, we chose to use the Binary Relevance (BR) method \cite{boutell2004learning}, a problem transformation approach in which we transform the problem into several binary datasets, one for each label. 
Though limited, in the sense that it assumes the labels are independent, this approach is simple and easily adapted to our problem. 
Thus, our meta-classifier is a set of \textit{m} classifiers, each one trained to indicate whether its corresponding model is relevant for a given input sample. 
However, since using the BR method may yield highly imbalanced binary problems, we train the meta-classifier applying class weights, with the weights being adjusted inversely proportional to the class frequencies in order to reduce the impact of the class imbalance.

\subsection{Generalization step}

\par
\begin{figure}
    \centering
    \includegraphics[width=0.5\textwidth]{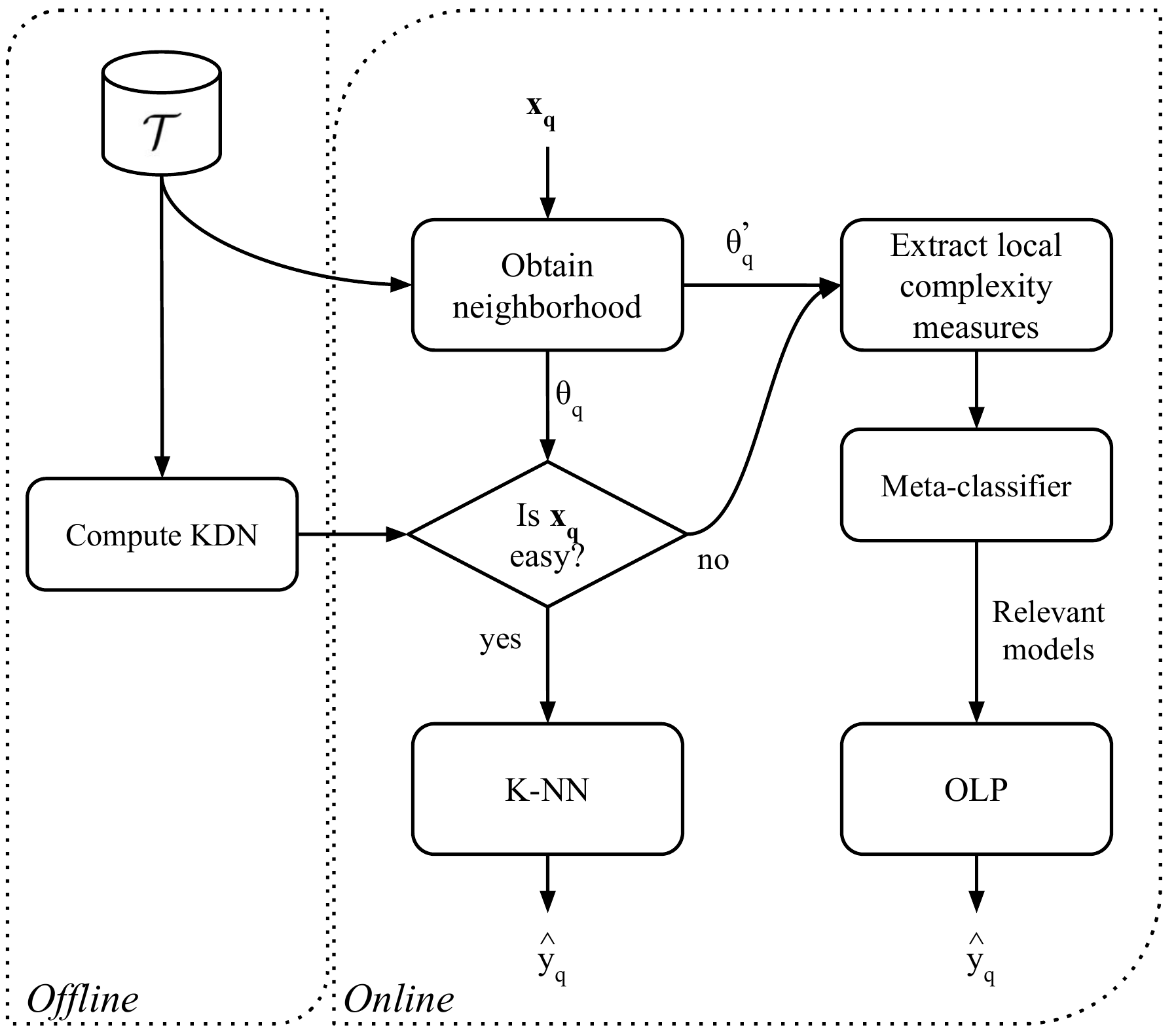}
    \caption{Overview of the generalization phase of the proposed framework.}
    \label{fig:overview-test}
\end{figure}

 In the generalization phase (Figure \ref{fig:overview-test}), the unseen training data is first analyzed and the training samples qualified into borderline (hard) or not (easy) using its KDN estimate. 
Then, in generalization, if the unknown sample $\mathbf{x_q}$ is considered easy, it is labelled by the K-NN, as in the original OLP technique. 
Otherwise, the local complexity measures are extracted over the neighborhood of the sample $\theta_q'$  with size $k'$, the same as in the meta-training step. 
The meta-feature vector $\mathbf{v_q}$ is then used as input to the meta-classifier, which returns which base-classifier models are relevant for the sample $\mathbf{x_q}$. 
Among the recommended models, the one whose meta-classifier (within the BR ensemble) outputs the highest class probability is chosen to be used.  
The OLP technique is then applied to the query sample with the chosen base-classifier model and yields the predicted label $\hat{y}_q$. 
If no model is recommended, the class probability rule is applied to all models (as in the T-Criterion rule for the BR method \cite{zhang2013review}).

\section{Experiments}
\label{sec:experimental-results}

\subsection{Experimental protocol}

\par The impact on the performance of the OLP using our meta-learning framework for choosing a relevant base-classifier model according to the local data complexity is assessed using a leave-one-dataset-out procedure, in order to achieve a problem-independent approach to our model recommendation scheme. 
That is, we use one dataset from our testbed in the generalization step and the remaining ones in the meta-training step for obtaining the meta-data. 
Within each dataset, we use a 5-fold cross validation procedure, both for algorithm evaluation and meta-feature extraction in the meta-training step and for evaluating the performance of the method in test in generalization. 

\paragraph{Datasets}

\par For facilitating the comparison with previous works, we use the same testbed of 64 two-class datasets from the Knowledge Extraction based on Evolutionary Learning (KEEL) repository \cite{keel}, presented in Table \ref{table:datasets}. 
Each dataset was evaluated using a stratified 5-fold cross validation procedure, one fold for test and the remaining for training, using the same partitions provided in the KEEL website for reproducibility. 
Due to the small-sized datasets, we use the training set as the DSEL set for the dynamic selection techniques evaluated, as in \cite{firedes++,souza2019oneval}.

\begin{table}[]
\setlength\tabcolsep{1.5pt}
\centering
\caption{Main characteristics of the datasets used in the experiments.}
\label{table:datasets}
\tiny
\centerline{
\scalebox{1}{
\begin{tabular}{@{}lllll|lllll@{}}
\toprule
\textbf{Ref.} & \textbf{Dataset}       & \textbf{\# Feat.} & \textbf{\# Samples} & \textbf{IR} & \textbf{Ref.} & \textbf{Dataset}             & \textbf{\# Feat.} & \textbf{\# Samples} & \textbf{IR} \\ \midrule
1             & glass1                 & 9                 & 214                 & 1.82        & 33            & ecoli-0-2-6-7vs3-5           & 7                 & 224                 & 9.18        \\
2             & ecoli0vs1              & 7                 & 220                 & 1.86        & 34            & glass-0-4vs5                 & 9                 & 92                  & 9.22        \\
3             & wisconsin              & 9                 & 683                 & 1.86        & 35            & ecoli-0-3-4-6vs5             & 7                 & 205                 & 9.25        \\
4             & pima                   & 8                 & 768                 & 1.87        & 36            & ecoli-0-3-4-7vs5-6           & 7                 & 257                 & 9.28        \\
5             & iris0                  & 4                 & 150                 & 2           & 37            & yeast-05679vs4               & 8                 & 528                 & 9.35        \\
6             & glass0                 & 9                 & 214                 & 2.06        & 38            & vowel0                       & 13                & 988                 & 9.98        \\
7             & yeast1                 & 8                 & 1484                & 2.46        & 39            & ecoli-0-6-7vs5               & 6                 & 220                 & 10          \\
8             & haberman               & 3                 & 306                 & 2.78        & 40            & glass-016vs2                 & 9                 & 192                 & 10.29       \\
9             & vehicle2               & 18                & 846                 & 2.88        & 41            & ecoli-0-1-4-7vs2-3-5-6       & 7                 & 336                 & 10.59       \\
10            & vehicle1               & 18                & 846                 & 2.9         & 42            & led7digit-0-2-4-5-6-7-8-9vs1 & 7                 & 443                 & 10.97       \\
11            & vehicle3               & 18                & 846                 & 2.99        & 43            & glass-0-6vs5                 & 9                 & 205                 & 11          \\
12            & glass0123vs456         & 9                 & 214                 & 3.2         & 44            & ecoli-0-1vs5                 & 6                 & 240                 & 11          \\
13            & vehicle0               & 18                & 846                 & 3.25        & 45            & glass-0-1-4-6vs2             & 9                 & 205                 & 11.06       \\
14            & ecoli1                 & 7                 & 336                 & 3.36        & 46            & glass2                       & 9                 & 214                 & 11.59       \\
15            & new-thyroid1           & 5                 & 215                 & 5.14        & 47            & ecoli-0-1-4-7vs5-6           & 6                 & 332                 & 12.28       \\
16            & new-thyroid2           & 5                 & 215                 & 5.14        & 48            & ecoli-0-1-4-6vs5             & 6                 & 280                 & 13          \\
17            & ecoli2                 & 7                 & 336                 & 5.46        & 49            & cleveland-0vs4               & 13                & 177                 & 12.62       \\
18            & segment0               & 19                & 2308                & 6           & 50            & shuttle-c0vsc4               & 9                 & 1829                & 13.87       \\
19            & glass6                 & 9                 & 214                 & 6.38        & 51            & yeast-1vs7                   & 7                 & 459                 & 14.3        \\
20            & yeast3                 & 8                 & 1484                & 8.1         & 52            & glass4                       & 9                 & 214                 & 15.47       \\
21            & ecoli3                 & 7                 & 336                 & 8.6         & 53            & ecoli4                       & 7                 & 336                 & 15.8        \\
22            & page-blocks0           & 10                & 5472                & 8.79        & 54            & page-blocks-13vs4            & 10                & 472                 & 15.86       \\
23            & ecoli-0-3-4vs5         & 7                 & 200                 & 9           & 55            & glass-0-1-6vs5               & 9                 & 184                 & 19.44       \\
24            & yeast-2vs4             & 8                 & 514                 & 9.08        & 56            & shuttle-c2-vs-c4             & 9                 & 129                 & 20.5        \\
25            & ecoli-0-6-7vs3-5       & 7                 & 202                 & 9.09        & 57            & yeast-1458vs7                & 8                 & 693                 & 22.1        \\
26            & ecoli-0-2-3-4vs5       & 7                 & 222                 & 9.1         & 58            & glass5                       & 9                 & 214                 & 22.78       \\
27            & yeast-0-3-5-9vs7-8     & 8                 & 506                 & 9.12        & 59            & yeast-2vs8                   & 8                 & 482                 & 23.1        \\
28            & glass-0-1-5vs2         & 9                 & 172                 & 9.12        & 60            & yeast4                       & 8                 & 1484                & 28.1        \\
29            & yeast-0-2-5-7-9vs3-6-8 & 8                 & 1004                & 9.14        & 61            & yeast-1289vs7                & 8                 & 947                 & 30.57       \\
30            & yeast-0-2-5-6vs3-7-8-9 & 8                 & 1004                & 9.14        & 62            & yeast5                       & 8                 & 1484                & 32.73       \\
31            & ecoli-0-4-6vs5         & 6                 & 203                 & 9.15        & 63            & ecoli-0137vs26               & 7                 & 281                 & 39.14       \\
32            & ecoli-0-1vs2-3-5       & 7                 & 224                 & 9.17        & 64            & yeast6                       & 8                 & 1484                & 41.4        \\ \bottomrule
\end{tabular}}}
\end{table}

\paragraph{Classifier models} 

\par We consider 5 base-classifier models to be chosen by the meta-learning framework: Perceptron, Decision Stump (DS), Decision Tree (DT), linear SVM (LSVM) and SVM with Gaussian kernel (GSVM). 
We compare the proposed framework with dynamic model type selection against the original OLP method, which generates the base-classifiers using the Self-Generating Hyperplanes (SGH) \cite{mariana} technique and uses a dynamic classifier selection technique embedded (we chose the Multiple Classifier Behavior (MCB) \cite{mcb} due to its superior performance in previous experiments). 
Moreover, we use the Decision Tree as our BR meta-learner in the multi-label framework. 
We chose this classifier because of its embedded feature selection, which allows us to analyze how correlated the local characteristics (meta-features) are with the relevance of each base-classifier model. 

\paragraph{Parameter setting} 

\par The pool size of the OLP, regardless of the base classifier used, is set to $M=5$. 
The neighborhood size for the KDN and the neighborhood definitions within the OLP framework are set to $k_h = k_s = 7$. 
For the version that uses the SGH and MCB, the similarity and competence threshold of the latter are set to $0.7$ and $0.1$, respectively. 

\par For the meta-learning framework, we fix the class probabilities threshold $t$ at $0.7$ in order to regard a base-classifier model as relevant or not for a given sample. 
Moreover, the neighborhood size for the meta-feature extraction is set to $k' = 50$, providing enough samples for reliably estimating the measures with a local scope \cite{kotsiantis_local_2006}. 
The meta-learner hyperparameters (maximum depth, minimum impurity decrease, minimum samples per leaf) are obtained using a grid search in a 10-fold cross-validation procedure over the meta-training set.

\paragraph{Performance measures}

\par In order to evaluate the impact of the automatic choice of base-classifier model on the performance of the OLP, we use the accuracy rate, the area under the Receiver Operating Characteristic (ROC) curve (AUC) \cite{auc}, the F-measure \cite{fm} and the Geometric Mean (G-mean) \cite{gm}, the latter three being measures frequently used for performance evaluation in imbalanced scenarios. 
Moreover, we use the Precision measure \cite{zhang2013review} for evaluating the performance of our multi-label meta-classifier, since for our scenario, it is more important that the set of recommended base-models is mostly comprised of suitable base-models than it contains a high proportion of unsuitable base-models, albeit including all suitable ones. 
In this sense, the precision of the meta-learner is the lower bound of our framework accuracy rate.

\subsection{Multi-label meta-classifier analysis}

\par
\begin{figure*}[]
    \centerline{
    \subfloat[Perceptron]{
    \includegraphics[width=0.34\textwidth]{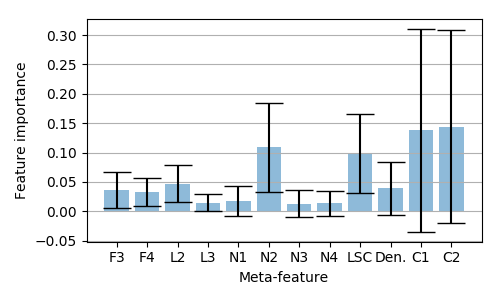}
    }
    \subfloat[DS]{
    \includegraphics[width=0.34\textwidth]{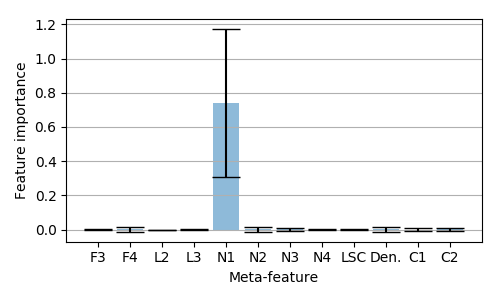}
    }
    \subfloat[DT]{
    \includegraphics[width=0.34\textwidth]{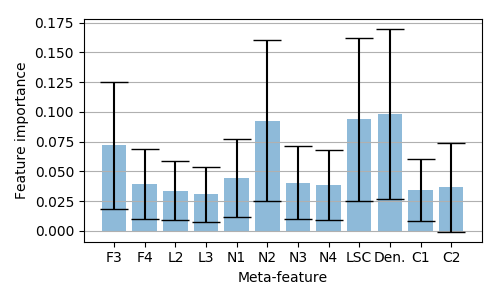}
    }
    }
    \centerline{
    \subfloat[LSVM]{
    \includegraphics[width=0.34\textwidth]{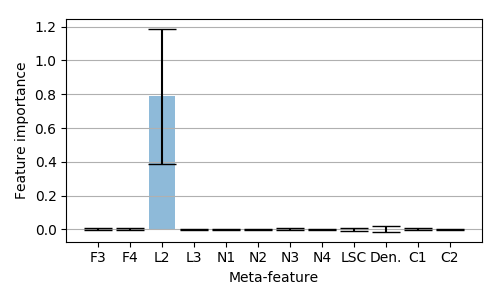}
    }
    \subfloat[GSVM]{
    \includegraphics[width=0.34\textwidth]{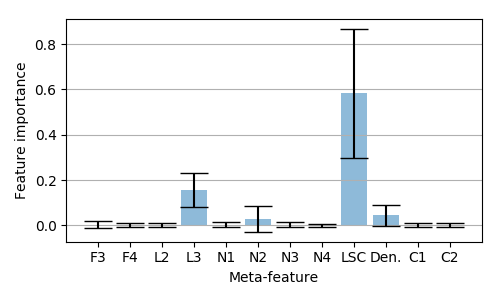}
    }
    }
    \caption{Mean feature importances of each component of the multi-label meta-classifier over all datasets from from Table \ref{table:datasets}.}
    \label{fig:feat-imp}
\end{figure*}

 We first analyze which characteristics (complexity measures) are more pertinent for recommending each base-classifier model. 
Figure \ref{fig:feat-imp} shows the mean meta-feature importances obtained from the multi-label meta-classifier (Decision tree). 
The importance of a feature in a DT is the normalized total reduction of the Gini impurity given by that feature. 
Since we used the BR method, the recommendation of each base-classifier model is given by an individual DT, so the feature importances for each model are shown in Figure \ref{fig:feat-imp}. 

\par Unsurprisingly, the most important meta-feature for recommending the LSVM is the L2 measure, which computes the error rate of a linear SVM on the local region. 
For the DS, the most important meta-feature appears to be the N1, which indicates the size of the local border. 
The recommendation of the Perceptron, on the other hand, was highly 
influenced by the local class balance and level of overlap in the target region. 
The level of class overlap and the linearity of the border is mostly regarded for recommending the GSVM. 
The recommendation of the DT, on the other hand, is quite different, with most meta-features being almost equally important. 

\begin{figure}[]
    \centering
    \subfloat[]{
    \includegraphics[width=0.37\textwidth]{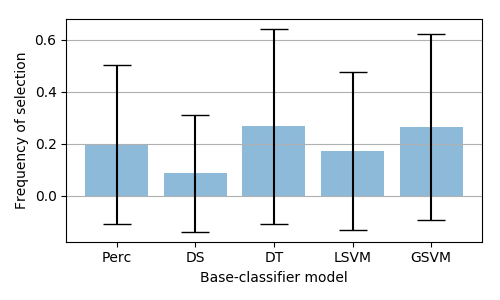}
    }
    \\
    \subfloat[]{
    \includegraphics[width=0.37\textwidth]{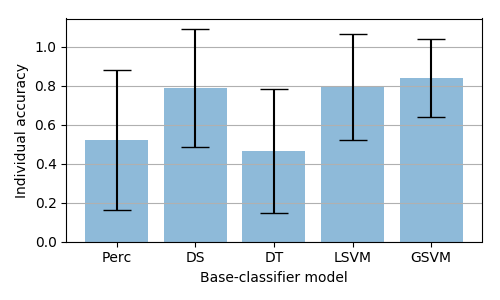}
    }
    \caption{(a) Mean frequency of selection of the base-classifier models and (b) mean individual accuracy of each component of the BR meta-classifier over all datasets from Table \ref{table:datasets}.}
    \label{fig:freq-acc}
\end{figure}

\par Figure \ref{fig:freq-acc}(a) shows the mean frequency of selection for each base-classifier model by the multi-label meta-classifier in generalization. 
It can be observed that the DS was overall selected less often, with the most frequently selected model being the DT. 
The individual mean accuracy rate of each component of the BR meta-classifier is shown in Figure \ref{fig:freq-acc}(b). 
It can be observed that, while the DT was the most recommended base-classifier model, its meta-classifier yielded the poorest accuracy rate, wrongly recommending the model half of the time, on average. 
The recommendation of the Perceptron was also generally quite inaccurate. 
For the remaining models, the mean accuracy rate was around $0.8$, which suggests that the meta-features used for characterizing their relevance in a subproblem are indeed important. 

\par
\begin{figure*}[]
\centerline{
    \includegraphics[width=0.67\textwidth]{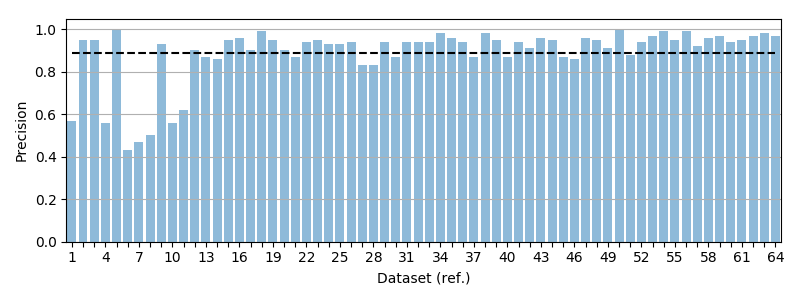}
    }
\caption{Mean precision of meta-classifier for each dataset in Table \ref{table:datasets}. 
The horizontal dashed line indicates the average performance over all datasets.}
    \label{fig:perf-metaclassif}
\end{figure*}

 The performance of the multi-label meta-classifier in generalization is depicted in Figure \ref{fig:perf-metaclassif}, which indicates the mean precision of the meta-classifier per dataset. 
It can be observed that the precision is quite high, especially for the highly imbalanced datasets (large reference number). 
However, for the first few datasets, the precision is quite low, reaching below $0.5$ for the \textit{glass0}, \textit{yeast1} and \textit{haberman} (ref. 6, 7 and 8) datasets. 
This may be explained by the labelset cardinality, that is, the average number of relevant classifiers per sample, which for these datasets, over which the meta-classifier yielded a poor precision score, is very low. 
This suggests that the multi-label meta-classifier struggles to recommend at least one relevant base-classifier model for the samples with low labelset cardinality. 

\paragraph{Framework performance}

\par We now analyze the performance of the framework as a whole. 
Table \ref{table:perf-all} shows the mean performance of the proposed framework (Proposed) and the online scheme using its original fixed individual base-classifier model (SGH+MCB) and an ideal base-classifier model for each sample. 
The results per dataset can be found in the Appendix. 
First, we can see that there is a significant improvement in selecting an ideal base-classifier model for each instance in particular in comparison to using the fixed model strategy for all instances, considering all performance metrics used in this work. 
Thus, we confirm one of our hypotheses: that each local region may favor certain types of classifiers and choosing the ideal one for each test sample is advantageous for local ensembles.
When we analyze the performance of the proposed technique, though, we see that the selection of such ideal base-classifier model per instance is not so trivial. 
In terms of accuracy, the proposed technique yielded a significantly inferior performance to using the fixed model strategy. 
However, in terms of AUC, F-measure and G-mean, the performance was statistically similar. 

\begin{table}[]
\setlength\tabcolsep{3pt}
\centering
\footnotesize
\caption{Average performance of the proposed framework (Proposed) and the online scheme using the fixed individual base-classifier model (SGH+MCB) and an ideal base-classifier model for each sample. 
Best results excluding the ideal selector ones are in bold. 
The row \textit{W-T-L} shows the number of wins, ties and losses of the proposed framework compared to using the column-wise strategy. 
The rows \textit{p-value} show the result of a Wilcoxon signed rank test with $\alpha=0.05$ between the indicated strategy (row-wise: proposed and ideal selector) and the column-wise strategy, with the symbols (+) and (-) indicating whether the former is significantly superior or inferior to the latter.}
\label{table:perf-all}
\begin{tabular}{@{}lllll@{}}
\toprule
\multicolumn{2}{l}{\textbf{Performance metric}}   & \textbf{Proposed} & \textbf{SGH+MCB}    &  \textbf{Ideal}     \\ \midrule
\multirow{4}{*}{Accuracy}  & Mean                 & 0.936             &  \textbf{0.941}     &  {\ul 0.971}        \\
                           & W-T-L                & n/a               &  14-20-30            &  0-5-59             \\
                           & p-value (proposed)   & n/a               &  \textbf{0.008 (-)} &  \textbf{0.000 (-)} \\
                           & p-value (ideal sel.) & -                 &  \textbf{0.000 (+)} &  n/a                \\ \midrule
\multirow{4}{*}{AUC}       & Mean                 & 0.805             &  \textbf{0.810}     &  {\ul 0.882}        \\
                           & W-T-L                & n/a               &  19-12-33           &  0-6-58            \\
                           & p-value (proposed)   & n/a               &  0.179              &  \textbf{0.000 (-)} \\
                           & p-value (ideal sel.) & -                 &  \textbf{0.000 (+)} &  n/a                \\ \midrule
\multirow{4}{*}{F-measure} & Mean                 & 0.674             &  \textbf{0.682}     &  {\ul 0.825}        \\
                           & W-T-L                & n/a               &  21-8-35           &  0-3-61             \\
                           & p-value (proposed)   & n/a               &  0.265              &  \textbf{0.000 (-)} \\
                           & p-value (ideal sel.) & -                 &  \textbf{0.000 (+)} &  n/a                \\ \midrule
\multirow{4}{*}{G-mean}    & Mean                 & \textbf{0.740}    &  \textbf{0.740}     &  {\ul 0.851}        \\
                           & W-T-L                & n/a               &  20-12-32           &  0-4-60             \\
                           & p-value (proposed)   & n/a               &  0.506              &  \textbf{0.000 (-)} \\
                           & p-value (ideal sel.) & -                 &  \textbf{0.000 (+)} &  n/a                \\ \bottomrule
\end{tabular}
\end{table}

\section{Conclusion}
\label{sec:conclusion}

\par In this work, we presented a novel algorithm recommendation framework which dynamically suggests a set of relevant model types for each instance in particular in a problem-independent manner. 
Since each model learns differently from a given set of points, our recommender system makes use of meta-learning and multi-label learning for recommending the models according to the local data complexity surrounding each test sample. 
We then integrated the algorithm recommender system to our online local pool generation technique \cite{souza2019online} and evaluated the proposed framework's performance over 64 binary problems.

\par Experiments showed that the local data characteristics affect the performance of each model type differently on a local scope. 
Moreover, it was shown that it is highly advantageous to use a suitable model type for each instance in particular, since the ideal model type selector yielded a statistically superior performance compared to the fixed model type approach considering all evaluated performance metrics. 
However, almost half of the components of our multi-label meta-classifier were not very well fit to the data, which may explain why its overall precision was high though it still struggled on harder recommendation scenarios. 
Overall, the performance of the proposed model type recommendation framework was statistically similar to using the original state-of-the-art online method, which uses a fixed base-classifier model for all test samples. 
Given the upper limit provided by the ideal selector's performance, and thus the margin for improvement of the framework, we believe this to be a promising line of research.

\par Since this is a novel approach to model type recommendation, in the sense that it is done dynamically according to the local structure of the data, there are many open challenges and improvements to be made. 
Future works may involve using a more powerful multi-label classifier that takes into account the label correlations of the meta-problem, as well as using a broader, more descriptive set of meta-features and applying a meta-feature selection procedure that is more suitable for multi-label learning.





\section*{Acknowledgments}

The authors would like to thank the Brazilian agencies CAPES (Coordena\c{c}\~{a}o de Aperfei\c{c}oamento de Pessoal de N\'{i}vel Superior), CNPq (Conselho Nacional de Desenvolvimento Cient\'{i}fico e Tecnol\'{o}gico) and FACEPE (Funda\c{c}\~{a}o de Amparo \`{a} Ci\^{e}ncia e Tecnologia de Pernambuco) and the Canadian agency NSERC (Natural Sciences and Engineering Research Council of Canada).

\bibliographystyle{IEEEtran}
\bibliography{references}

\appendix[Detailed performance results]

\begin{table}[H]
\setlength\tabcolsep{3pt}
\caption{Mean and standard deviation of the accuracy rate of the evaluated techniques over each dataset from Table \ref{table:datasets}. 
Best results are in bold.}
\label{table:acc-all}
\scriptsize
\centerline{
\scalebox{0.88}{
\begin{tabular}{@{}llll|llll@{}}
\toprule
\textbf{Ref.} & \textbf{Proposed} & \textbf{SGH+MCB} & \textbf{Ideal} & \textbf{Ref.} & \textbf{Proposed} & \textbf{SGH+MCB} & \textbf{Ideal} \\ \midrule
1             & 0.76 (0.08)      &  \textbf{0.77} (0.07)   & 0.91 (0.05)  & 33            & \textbf{0.96} (0.01)               & \textbf{0.96} (0.01)  & 0.98 (0.01)  \\
2             & \textbf{0.98} (0.01)      &  0.97 (0.02)   & 0.99 (0.01)  & 34            & \textbf{1.00} (0.00)               & 0.99 (0.02)  & 1.00 (0.00)  \\
3             & 0.96 (0.01)      &  \textbf{0.97} (0.01)   & 0.98 (0.01)  & 35            & \textbf{0.97} (0.02)               & \textbf{0.97} (0.02)  & 0.99 (0.01)  \\
4             & 0.73 (0.02)      &  \textbf{0.76} (0.04)   & 0.86 (0.02)  & 36            & \textbf{0.97} (0.02)               & 0.96 (0.02)  & 0.98 (0.03)  \\
5             & \textbf{1.00} (0.00)      &  \textbf{1.00} (0.00)   & 1.00 (0.00)  & 37            & \textbf{0.91} (0.01)               & \textbf{0.91} (0.01)  & 0.95 (0.01)  \\
6             & 0.79 (0.02)      &  \textbf{0.84} (0.06)   & 0.94 (0.04)  & 38            & \textbf{0.99} (0.01)               & \textbf{0.99} (0.00)  & 1.00 (0.00)  \\
7             & 0.72 (0.02)      &  \textbf{0.75} (0.03)   & 0.90 (0.01)  & 39            & \textbf{0.97} (0.01)               & \textbf{0.97} (0.02)  & 0.98 (0.02)  \\
8             & 0.69 (0.03)      &  \textbf{0.71} (0.05)   & 0.87 (0.03)  & 40            & 0.90 (0.02)               & \textbf{0.91} (0.03)  & 0.94 (0.03)  \\
9             & \textbf{0.98} (0.01)      &  0.97 (0.01)   & 0.99 (0.00)  & 41            & 0.95 (0.02)               & \textbf{0.97} (0.02)  & 0.98 (0.01)  \\
10            & 0.75 (0.01)      &  \textbf{0.79} (0.02)   & 0.92 (0.02)  & 42            & \textbf{0.97} (0.01)               & 0.95 (0.01)  & 0.97 (0.02)  \\
11            & 0.79 (0.01)      &  \textbf{0.80} (0.02)   & 0.93 (0.01)  & 43            & \textbf{0.99} (0.01)               & \textbf{0.99} (0.02)  & 0.99 (0.02)  \\
12            & \textbf{0.93} (0.02)      &  \textbf{0.93} (0.01)   & 0.98 (0.01)  & 44            & \textbf{0.98} (0.01)               & 0.97 (0.02)  & 0.98 (0.02)  \\
13            & 0.95 (0.02)      &  \textbf{0.96} (0.02)   & 1.00 (0.01)  & 45            & 0.90 (0.01)               & \textbf{0.92} (0.02)  & 0.96 (0.02)  \\
14            & 0.91 (0.02)      &  \textbf{0.92} (0.03)   & 0.96 (0.02)  & 46            & \textbf{0.91} (0.03)               & 0.90 (0.02)  & 0.95 (0.03)  \\
15            & 0.98 (0.01)      &  \textbf{0.99} (0.01)   & 1.00 (0.01)  & 47            & \textbf{0.98} (0.01)               & 0.97 (0.02)  & 0.99 (0.01)  \\
16            & 0.98 (0.01)      &  \textbf{0.99} (0.02)   & 1.00 (0.00)  & 48            & \textbf{0.98} (0.01)               & 0.97 (0.02)  & 0.99 (0.01)  \\
17            & 0.95 (0.02)      &  \textbf{0.96} (0.03)   & 0.98 (0.02)  & 49            & 0.93 (0.02)               & \textbf{0.94} (0.02)  & 0.98 (0.02)  \\
18            & \textbf{0.99} (0.00)      &  \textbf{0.99} (0.00)   & 1.00 (0.00)  & 50            & \textbf{1.00} (0.00)               & \textbf{1.00} (0.00)  & 1.00 (0.00)  \\
19            & \textbf{0.97} (0.01)      &  0.96 (0.01)   & 0.97 (0.02)  & 51            & 0.93 (0.01)               & \textbf{0.94} (0.01)  & 0.96 (0.01)  \\
20            & 0.93 (0.01)      &  \textbf{0.95} (0.01)   & 0.97 (0.01)  & 52            & 0.96 (0.03)               & \textbf{0.97} (0.03)  & 0.98 (0.02)  \\
21            & 0.90 (0.01)      &  \textbf{0.92} (0.03)   & 0.96 (0.02)  & 53            & 0.98 (0.01)               & \textbf{0.99} (0.01)  & 0.99 (0.01)  \\
22            & \textbf{0.97} (0.01)      &  \textbf{0.97} (0.00)   & 0.99 (0.00)  & 54            & \textbf{1.00} (0.00)               & 0.99 (0.01)  & 1.00 (0.00)  \\
23            & \textbf{0.97} (0.02)      &  \textbf{0.97} (0.03)   & 0.97 (0.02)  & 55            & \textbf{0.98} (0.02)               & 0.97 (0.03)  & 0.99 (0.01)  \\
24            & 0.94 (0.01)      &  \textbf{0.96} (0.01)   & 0.98 (0.02)  & 56            & \textbf{0.99} (0.01)               & \textbf{0.99} (0.02)  & 0.99 (0.02)  \\
25            & \textbf{0.96} (0.03)      &  0.94 (0.03)   & 0.98 (0.03)  & 57            & 0.94 (0.01)               & \textbf{0.95} (0.01)  & 0.96 (0.01)  \\
26            & \textbf{0.97} (0.02)      &  \textbf{0.97} (0.03)   & 0.98 (0.02)  & 58            & \textbf{0.98} (0.02)               & \textbf{0.98} (0.02)  & 1.00 (0.01)  \\
27            & \textbf{0.91} (0.01)      &  0.90 (0.01)   & 0.95 (0.01)  & 59            & 0.97 (0.01)               & \textbf{0.98} (0.01)  & 0.98 (0.01)  \\
28            & \textbf{0.87} (0.03)      &  \textbf{0.87} (0.07)   & 0.92 (0.03)  & 60            & 0.96 (0.01)               & \textbf{0.97} (0.01)  & 0.98 (0.01)  \\
29            & 0.96 (0.01)      &  \textbf{0.97} (0.01)   & 0.98 (0.01)  & 61            & 0.96 (0.01)               & \textbf{0.97} (0.01)  & 0.98 (0.00)  \\
30            & \textbf{0.93} (0.01)      &  \textbf{0.93} (0.02)   & 0.96 (0.01)  & 62            & \textbf{0.98} (0.01)               & \textbf{0.98} (0.00)  & 0.99 (0.00)  \\
31            & 0.96 (0.02)      &  \textbf{0.97} (0.04)   & 0.99 (0.02)  & 63            & \textbf{0.99} (0.01)               & \textbf{0.99} (0.01)  & 0.99 (0.01)  \\
32            & 0.95 (0.02)      &  \textbf{0.97} (0.03)   & 0.98 (0.02)  & 64            & \textbf{0.98} (0.01)               & \textbf{0.98} (0.00)  & 0.99 (0.01)  \\ \bottomrule
\end{tabular}
}
}
\end{table}

\begin{table}[H]
\setlength\tabcolsep{3pt}
\caption{Mean and standard deviation of the AUC of the evaluated techniques over each dataset from Table \ref{table:datasets}. 
Best results are in bold.}
\label{table:auc-all}
\scriptsize
\centerline{
\scalebox{0.88}{
\begin{tabular}{@{}llll|llll@{}}
\toprule
\textbf{Ref.} & \textbf{Proposed} & \textbf{SGH+MCB} & \textbf{Ideal} & \textbf{Ref.} & \textbf{Proposed} & \textbf{SGH+MCB} & \textbf{Ideal} \\ \midrule
1             & 0.73 (0.09)  &  \textbf{0.74} (0.08)    &  0.89 (0.06)  & 33            &   0.83 (0.09)  &  \textbf{0.85} (0.09)   &  0.90 (0.09)  \\
2             & \textbf{0.97} (0.02)  &  0.96 (0.03)    &  0.99 (0.02)  & 34            &   \textbf{1.00} (0.00)  &  0.95 (0.10)   &  1.00 (0.00)  \\
3             & 0.95 (0.01)  &  \textbf{0.97} (0.01)    &  0.98 (0.01)  & 35            &   \textbf{0.89} (0.09)  &  \textbf{0.89} (0.09)   &  0.93 (0.06)  \\
4             & 0.69 (0.04)  &  \textbf{0.74} (0.04)    &  0.84 (0.02)  & 36            &   \textbf{0.89} (0.11)  &  \textbf{0.89} (0.09)   &  0.92 (0.12)  \\
5             & \textbf{1.00} (0.00)  &  \textbf{1.00} (0.00)    &  1.00 (0.00)  & 37            &   \textbf{0.67} (0.08)  &  0.63 (0.09)   &  0.81 (0.05)  \\
6             & 0.77 (0.03)  &  \textbf{0.82} (0.08)    &  0.94 (0.05)  & 38            &   0.95 (0.04)  &  \textbf{0.98} (0.01)   &  1.00 (0.00)  \\
7             & 0.67 (0.04)  &  \textbf{0.69} (0.04)    &  0.86 (0.02)  & 39            &   \textbf{0.87} (0.07)  &  \textbf{0.87} (0.08)   &  0.92 (0.06)  \\
8             & 0.57 (0.05)  &  \textbf{0.58} (0.05)    &  0.79 (0.05)  & 40            &   0.52 (0.07)  &  \textbf{0.53} (0.07)   &  0.68 (0.13)  \\
9             & \textbf{0.96} (0.01)  &  0.95 (0.01)    &  0.99 (0.01)  & 41            &   0.80 (0.05)  &  \textbf{0.87} (0.08)   &  0.90 (0.06)  \\
10            & 0.66 (0.03)  &  \textbf{0.68} (0.01)    &  0.87 (0.03)  & 42            &   \textbf{0.88} (0.06)  &  0.78 (0.06)   &  0.90 (0.08)  \\
11            & 0.70 (0.03)  &  \textbf{0.73} (0.02)    &  0.89 (0.02)  & 43            &   \textbf{0.95} (0.10)  &  \textbf{0.95} (0.10)   &  0.95 (0.10)  \\
12            & 0.88 (0.06)  &  \textbf{0.90} (0.03)    &  0.96 (0.03)  & 44            &   \textbf{0.90} (0.09)  &  0.89 (0.09)   &  0.90 (0.09)  \\
13            & \textbf{0.94} (0.03)  &  \textbf{0.94} (0.02)    &  0.99 (0.01)  & 45            &   \textbf{0.55} (0.06)  &  0.54 (0.10)   &  0.72 (0.14)  \\
14            & \textbf{0.87} (0.04)  &  \textbf{0.87} (0.04)    &  0.94 (0.04)  & 46            &   \textbf{0.55} (0.09)  &  0.49 (0.01)   &  0.69 (0.13)  \\
15            & 0.95 (0.03)  &  \textbf{0.97} (0.03)    &  0.99 (0.03)  & 47            &   \textbf{0.88} (0.07)  &  0.86 (0.08)   &  0.92 (0.04)  \\
16            & 0.94 (0.05)  &  \textbf{0.97} (0.06)    &  1.00 (0.00)  & 48            &   \textbf{0.90} (0.12)  &  0.87 (0.15)   &  0.93 (0.10)  \\
17            & 0.87 (0.04)  &  \textbf{0.90} (0.03)    &  0.94 (0.04)  & 49            &   0.58 (0.11)  &  \textbf{0.68} (0.11)   &  0.85 (0.13)  \\
18            & \textbf{0.99} (0.01)  &  0.98 (0.01)    &  0.99 (0.01)  & 50            &   \textbf{1.00} (0.01)  &  \textbf{1.00} (0.01)   &  1.00 (0.01)  \\
19            & \textbf{0.89} (0.07)  &  \textbf{0.89} (0.04)    &  0.91 (0.06)  & 51            &   0.62 (0.03)  &  \textbf{0.64} (0.06)   &  0.70 (0.08)  \\
20            & \textbf{0.82} (0.01)  &  \textbf{0.82} (0.02)    &  0.91 (0.03)  & 52            &   0.74 (0.13)  &  \textbf{0.84} (0.13)   &  0.88 (0.14)  \\
21            & 0.73 (0.07)  &  \textbf{0.77} (0.10)    &  0.86 (0.09)  & 53            &   0.89 (0.04)  &  \textbf{0.90} (0.05)   &  0.90 (0.05)  \\
22            & \textbf{0.92} (0.01)  &  \textbf{0.92} (0.01)    &  0.96 (0.01)  & 54            &   \textbf{0.97} (0.04)  &  0.95 (0.07)   &  1.00 (0.00)  \\
23            & 0.87 (0.12)  &  \textbf{0.89} (0.10)    &  0.88 (0.11)  & 55            &   \textbf{0.85} (0.20)  &  0.79 (0.19)   &  0.95 (0.10)  \\
24            & 0.82 (0.03)  &  \textbf{0.87} (0.03)    &  0.92 (0.06)  & 56            &   \textbf{0.95} (0.10)  &  \textbf{0.95} (0.10)   &  0.95 (0.10)  \\
25            & \textbf{0.86} (0.17)  &  0.80 (0.16)    &  0.92 (0.12)  & 57            &   \textbf{0.51} (0.03)  &  0.49 (0.00)   &  0.55 (0.07)  \\
26            & 0.85 (0.09)  &  \textbf{0.87} (0.12)    &  0.88 (0.11)  & 58            &   \textbf{0.80} (0.24)  &  0.75 (0.22)   &  0.95 (0.10)  \\
27            & 0.60 (0.04)  &  \textbf{0.61} (0.04)    &  0.75 (0.05)  & 59            &   0.70 (0.05)  &  \textbf{0.74} (0.10)   &  0.77 (0.09)  \\
28            & \textbf{0.54} (0.08)  &  0.48 (0.03)    &  0.60 (0.13)  & 60            &   0.61 (0.07)  &  \textbf{0.63} (0.07)   &  0.73 (0.08)  \\
29            & 0.88 (0.04)  &  \textbf{0.89} (0.03)    &  0.91 (0.03)  & 61            &   0.59 (0.03)  &  \textbf{0.63} (0.08)   &  0.65 (0.06)  \\
30            & 0.72 (0.03)  &  \textbf{0.75} (0.04)    &  0.81 (0.04)  & 62            &   \textbf{0.84} (0.09)  &  0.74 (0.09)   &  0.94 (0.04)  \\
31            & 0.87 (0.13)  &  \textbf{0.89} (0.15)    &  0.93 (0.10)  & 63            &   0.80 (0.19)  &  \textbf{0.85} (0.20)   &  0.85 (0.20)  \\
32            & 0.80 (0.10)  &  \textbf{0.86} (0.14)    &  0.86 (0.14)  & 64            &   \textbf{0.73} (0.13)  &  0.67 (0.07)   &  0.83 (0.11)  \\ \bottomrule
\end{tabular}
}
}
\end{table}

\rule{0pt}{14ex} 

\begin{table}[H]
\setlength\tabcolsep{3pt}
\caption{Mean and standard deviation of the F-measure of the evaluated techniques over each dataset from Table \ref{table:datasets}. 
Best results are in bold.}
\label{table:fm-all}
\scriptsize
\centerline{
\scalebox{0.88}{
\begin{tabular}{@{}llll|llll@{}}
\toprule
\textbf{Ref.} & \textbf{Proposed} & \textbf{SGH+MCB} & \textbf{Ideal} 											& \textbf{Ref.} & \textbf{Proposed} & \textbf{SGH+MCB} & \textbf{Ideal} \\ \midrule
1             & 0.64 (0.12)  & \textbf{0.66} (0.12)  & 0.87 (0.08) & 33            &   0.75 (0.12)  &  \textbf{0.78} (0.10)  &  0.86 (0.11)  \\
2             & 0.97 (0.03)  & \textbf{0.98} (0.01)  & 0.99 (0.01) & 34            &   \textbf{1.00} (0.00)  &  0.93 (0.13)  &  1.00 (0.00)  \\
3             & 0.94 (0.01)  & \textbf{0.96} (0.01)  & 0.97 (0.01) & 35            &   \textbf{0.84} (0.15)  &  0.81 (0.12)  &  0.91 (0.07)  \\
4             & 0.60 (0.06)  & \textbf{0.65} (0.06)  & 0.79 (0.03) & 36            &   0.80 (0.17)  &  \textbf{0.81} (0.13)  &  0.88 (0.19)  \\
5             & \textbf{1.00} (0.00)  & \textbf{1.00} (0.00)  & 1.00 (0.00) & 37            &   \textbf{0.42} (0.16)  &  0.34 (0.20)  &  0.71 (0.06)  \\
6             & 0.69 (0.04)  & \textbf{0.75} (0.11)  & 0.91 (0.06) & 38            &   0.93 (0.07)  &  \textbf{0.97} (0.01)  &  1.00 (0.00)  \\
7             & 0.53 (0.06)  & \textbf{0.56} (0.05)  & 0.81 (0.03) & 39            &   \textbf{0.83} (0.11)  &  \textbf{0.83} (0.11)  &  0.89 (0.10)  \\
8             & 0.35 (0.09)  & \textbf{0.36} (0.09)  & 0.70 (0.08) & 40            &   \textbf{0.10} (0.20)  &  \textbf{0.10} (0.20)  &  0.48 (0.30)  \\
9             & \textbf{0.95} (0.02)  & 0.94 (0.02)  & 0.99 (0.01) & 41            &   0.67 (0.08)  &  \textbf{0.81} (0.13)  &  0.87 (0.05)  \\
10            & 0.50 (0.05)  & \textbf{0.53} (0.02)  & 0.82 (0.05) & 42            &   \textbf{0.79} (0.10)  &  0.63 (0.10)  &  0.80 (0.11)  \\
11            & 0.56 (0.05)  & \textbf{0.60} (0.03)  & 0.85 (0.02) & 43            &   \textbf{0.93} (0.13)  &  \textbf{0.93} (0.13)  &  0.93 (0.13)  \\
12            & 0.83 (0.07)  & \textbf{0.85} (0.02)  & 0.95 (0.04) & 44            &   \textbf{0.85} (0.11)  &  0.81 (0.12)  &  0.88 (0.12)  \\
13            & 0.90 (0.05)  & \textbf{0.92} (0.03)  & 0.99 (0.01) & 45            &   \textbf{0.14} (0.17)  &  0.11 (0.23)  &  0.54 (0.31)  \\
14            & 0.81 (0.05)  & \textbf{0.82} (0.07)  & 0.92 (0.05) & 46            &   \textbf{0.20} (0.24)  &  0.00 (0.00)  &  0.50 (0.30)  \\
15            & 0.94 (0.03)  & \textbf{0.96} (0.04)  & 0.98 (0.03) & 47            &   \textbf{0.82} (0.11)  &  0.78 (0.12)  &  0.91 (0.04)  \\
16            & 0.94 (0.06)  & \textbf{0.97} (0.07)  & 1.00 (0.00) & 48            &   \textbf{0.84} (0.15)  &  0.75 (0.21)  &  0.90 (0.13)  \\
17            & 0.81 (0.07)  & \textbf{0.85} (0.08)  & 0.92 (0.05) & 49            &   0.23 (0.29)  &  \textbf{0.46} (0.26)  &  0.79 (0.19)  \\
18            & \textbf{0.98} (0.01)  & \textbf{0.98} (0.01)  & 0.99 (0.01) & 50            &   \textbf{1.00} (0.01)  &  \textbf{1.00} (0.01)  &  1.00 (0.01)  \\
19            & \textbf{0.86} (0.09)  & 0.85 (0.06)  & 0.88 (0.08) & 51            &   0.32 (0.06)  &  \textbf{0.38} (0.13)  &  0.55 (0.17)  \\
20            & 0.69 (0.02)  & \textbf{0.73} (0.02)  & 0.87 (0.04) & 52            &   0.60 (0.23)  &  \textbf{0.75} (0.21)  &  0.80 (0.20)  \\
21            & 0.51 (0.10)  & \textbf{0.59} (0.18)  & 0.79 (0.13) & 53            &   0.80 (0.05)  &  \textbf{0.86} (0.08)  &  0.89 (0.06)  \\
22            & \textbf{0.85} (0.02)  & 0.84 (0.01)  & 0.94 (0.01) & 54            &   \textbf{0.96} (0.04)  &  0.94 (0.08)  &  1.00 (0.00)  \\
23            & 0.82 (0.17)  & \textbf{0.84} (0.16)  & 0.84 (0.15) & 55            &   \textbf{0.69} (0.37)  &  0.60 (0.37)  &  0.93 (0.13)  \\
24            & 0.69 (0.02)  & \textbf{0.79} (0.03)  & 0.88 (0.09) & 56            &   \textbf{0.93} (0.13)  &  \textbf{0.93} (0.13)  &  0.93 (0.13)  \\
25            & \textbf{0.76} (0.26)  & 0.64 (0.21)  & 0.89 (0.17) & 57            &   \textbf{0.06} (0.10)  &  0.00 (0.00)  &  0.16 (0.20)  \\
26            & 0.79 (0.15)  & \textbf{0.82} (0.17)  & 0.84 (0.15) & 58            &   \textbf{0.60} (0.49)  &  0.53 (0.45)  &  0.93 (0.13)  \\
27            & \textbf{0.31} (0.10)  & \textbf{0.31} (0.08)  & 0.65 (0.10) & 59            &   0.51 (0.10)  &  \textbf{0.65} (0.17)  &  0.67 (0.15)  \\
28            & \textbf{0.15} (0.18)  & 0.00 (0.00)  & 0.26 (0.33) & 60            &   0.27 (0.15)  &  \textbf{0.36} (0.16)  &  0.61 (0.16)  \\
29            & 0.79 (0.05)  & \textbf{0.83} (0.06)  & 0.89 (0.05) & 61            &   0.25 (0.06)  &  \textbf{0.34} (0.20)  &  0.45 (0.15)  \\
30            & 0.55 (0.07)  & \textbf{0.60} (0.08)  & 0.74 (0.05) & 62            &   \textbf{0.69} (0.15)  &  0.57 (0.14)  &  0.90 (0.04)  \\
31            & 0.76 (0.20)  & \textbf{0.82} (0.25)  & 0.90 (0.13) & 63            &   \textbf{0.67} (0.37)  &  0.63 (0.37)  &  0.73 (0.39)  \\
32            & 0.67 (0.18)  & \textbf{0.77} (0.23)  & 0.81 (0.22) & 64            &   \textbf{0.51} (0.26)  &  0.45 (0.14)  &  0.76 (0.15)  \\ \bottomrule
\end{tabular}
}
}
\end{table}

\begin{table}[H]
\setlength\tabcolsep{3pt}
\caption{Mean and standard deviation of the geometric mean of the evaluated techniques over each dataset from Table \ref{table:datasets}. 
Best results are in bold.}
\label{table:gm-all}
\scriptsize
\centerline{
\scalebox{0.88}{
\begin{tabular}{@{}llll|llll@{}}
\toprule
\textbf{Ref.} & \textbf{Proposed} & \textbf{SGH+MCB} & \textbf{Ideal} & \textbf{Ref.} & \textbf{Proposed} & \textbf{SGH+MCB} & \textbf{Ideal} \\ \midrule
1             & 0.72 (0.10)   &  \textbf{0.73} (0.09)  & 0.89 (0.07)  & 33            &  0.81 (0.11)   & \textbf{0.83} (0.11) &  0.89 (0.11) \\
2             & \textbf{0.97} (0.03)   &  0.96 (0.03)  & 0.99 (0.02)  & 34            &  \textbf{1.00} (0.00)   & 0.94 (0.12) &  1.00 (0.00) \\
3             & 0.95 (0.01)   &  \textbf{0.97} (0.01)  & 0.98 (0.01)  & 35            &  \textbf{0.88} (0.11)   & \textbf{0.88} (0.11) &  0.92 (0.07) \\
4             & 0.68 (0.05)   &  \textbf{0.73} (0.05)  & 0.83 (0.02)  & 36            &  \textbf{0.88} (0.13)   & \textbf{0.88} (0.10) &  0.90 (0.15) \\
5             & \textbf{1.00} (0.00)   &  \textbf{1.00} (0.00)  & 1.00 (0.00)  & 37            &  \textbf{0.58} (0.15)   & 0.45 (0.25) &  0.78 (0.06) \\
6             & 0.77 (0.04)   &  \textbf{0.81} (0.09)  & 0.94 (0.05)  & 38            &  0.95 (0.04)   & \textbf{0.98} (0.01) &  1.00 (0.00) \\
7             & 0.65 (0.05)   &  \textbf{0.67} (0.04)  & 0.85 (0.02)  & 39            &  \textbf{0.86} (0.09)   & \textbf{0.86} (0.09) &  0.92 (0.07) \\
8             & \textbf{0.51} (0.07)   &  \textbf{0.51} (0.07)  & 0.76 (0.07)  & 40            &  \textbf{0.12} (0.23)   & \textbf{0.12} (0.23) &  0.53 (0.30) \\
9             & \textbf{0.96} (0.01)   &  0.95 (0.01)  & 0.99 (0.01)  & 41            &  0.78 (0.07)   & \textbf{0.86} (0.08) &  0.89 (0.07) \\
10            & 0.63 (0.04)   &  \textbf{0.64} (0.01)  & 0.86 (0.03)  & 42            &  \textbf{0.87} (0.07)   & 0.74 (0.08) &  0.89 (0.09) \\
11            & 0.68 (0.05)   &  \textbf{0.72} (0.02)  & 0.89 (0.02)  & 43            &  \textbf{0.94} (0.12)   & \textbf{0.94} (0.12) &  0.94 (0.12) \\
12            & 0.87 (0.07)   &  \textbf{0.90} (0.03)  & 0.96 (0.04)  & 44            &  \textbf{0.89} (0.11)   & 0.88 (0.11) &  0.89 (0.11) \\
13            & 0.93 (0.03)   &  \textbf{0.94} (0.02)  & 0.99 (0.01)  & 45            &  \textbf{0.21} (0.26)   & 0.14 (0.28) &  0.58 (0.32) \\
14            & \textbf{0.87} (0.04)   &  \textbf{0.87} (0.04)  & 0.94 (0.04)  & 46            &  \textbf{0.23} (0.28)   & 0.00 (0.00) &  0.54 (0.30) \\
15            & 0.95 (0.03)   &  \textbf{0.97} (0.03)  & 0.99 (0.03)  & 47            &  \textbf{0.86} (0.09)   & 0.84 (0.09) &  0.92 (0.04) \\
16            & 0.94 (0.06)   &  \textbf{0.97} (0.06)  & 1.00 (0.00)  & 48            &  \textbf{0.88} (0.14)   & 0.84 (0.20) &  0.91 (0.12) \\
17            & 0.86 (0.04)   &  \textbf{0.89} (0.04)  & 0.94 (0.04)  & 49            &  0.26 (0.32)   & \textbf{0.53} (0.28) &  0.82 (0.17) \\
18            & \textbf{0.99} (0.01)   &  0.98 (0.01)  & 0.99 (0.01)  & 50            &  \textbf{1.00} (0.01)   & \textbf{1.00} (0.01) &  1.00 (0.01) \\
19            & \textbf{0.88} (0.08)   &  \textbf{0.88} (0.05)  & 0.90 (0.07)  & 51            &  0.50 (0.08)   & \textbf{0.53} (0.11) &  0.62 (0.14) \\
20            & \textbf{0.81} (0.01)   &  0.80 (0.02)  & 0.90 (0.04)  & 52            &  0.68 (0.17)   & \textbf{0.82} (0.17) &  0.85 (0.18) \\
21            & 0.69 (0.10)   &  \textbf{0.73} (0.14)  & 0.85 (0.11)  & 53            &  \textbf{0.89} (0.05)   & \textbf{0.89} (0.05) &  0.89 (0.05) \\
22            & 0.91 (0.01)   &  \textbf{0.92} (0.01)  & 0.96 (0.01)  & 54            &  \textbf{0.97} (0.04)   & 0.94 (0.07) &  1.00 (0.00) \\
23            & 0.85 (0.13)   &  \textbf{0.88} (0.11)  & 0.86 (0.13)  & 55            &  \textbf{0.74} (0.39)   & 0.68 (0.37) &  0.94 (0.12) \\
24            & 0.80 (0.05)   &  \textbf{0.86} (0.04)  & 0.91 (0.07)  & 56            &  \textbf{0.94} (0.12)   & \textbf{0.94} (0.12) &  0.94 (0.12) \\
25            & \textbf{0.81} (0.23)   &  0.75 (0.21)  & 0.91 (0.14)  & 57            &  \textbf{0.08} (0.16)   & 0.00 (0.00) &  0.20 (0.25) \\
26            & 0.83 (0.11)   &  \textbf{0.85} (0.13)  & 0.86 (0.13)  & 58            &  \textbf{0.60} (0.49)   & 0.54 (0.45) &  0.94 (0.12) \\
27            & 0.45 (0.10)   &  \textbf{0.47} (0.10)  & 0.70 (0.08)  & 59            &  0.62 (0.10)   & \textbf{0.70} (0.14) &  0.73 (0.13) \\
28            & \textbf{0.23} (0.28)   &  0.00 (0.00)  & 0.28 (0.35)  & 60            &  0.46 (0.14)   & \textbf{0.49} (0.13) &  0.66 (0.12) \\
29            & 0.87 (0.05)   &  \textbf{0.88} (0.04)  & 0.90 (0.04)  & 61            &  0.44 (0.07)   & \textbf{0.45} (0.25) &  0.54 (0.11) \\
30            & 0.66 (0.05)   &  \textbf{0.71} (0.05)  & 0.79 (0.05)  & 62            &  \textbf{0.81} (0.11)   & 0.69 (0.13) &  0.94 (0.04) \\
31            & 0.84 (0.18)   &  \textbf{0.87} (0.19)  & 0.91 (0.12)  & 63            &  0.68 (0.37)   & \textbf{0.74} (0.39) &  0.74 (0.39) \\
32            & 0.76 (0.14)   &  \textbf{0.83} (0.19)  & 0.83 (0.19)  & 64            &  \textbf{0.64} (0.22)   & 0.57 (0.13) &  0.80 (0.13) \\ \bottomrule
\end{tabular}
}
}
\end{table}

\end{document}